\documentclass[11pt,a4paper]{article}
\usepackage[hyperref]{emnlp-ijcnlp-2019}
\usepackage{times}
\usepackage{latexsym}
\usepackage{url}
\usepackage{multirow}
\usepackage{adjustbox}
\usepackage{amssymb,amsmath}

\usepackage{algorithm}      % algorithm
\usepackage{algpseudocode}
\usepackage{bm}             % bold in math
\usepackage{graphicx}  %Required
\usepackage{subfigure}
\usepackage{amsfonts}
\usepackage{multirow}
\usepackage{makecell}

\aclfinalcopy 

\title{CM-Net: A Novel Collaborative Memory Network for \\Spoken Language Understanding}

\author{
  Yijin Liu\textsuperscript{1}\thanks{\ \ This work was done when Yijin Liu was interning at Pattern Recognition Center, WeChat AI, Tencent Inc, China} \ ,
  Fandong Meng\textsuperscript{2}, 
  Jinchao Zhang\textsuperscript{2}, 
  Jie Zhou\textsuperscript{2},
  Yufeng Chen\textsuperscript{1} 
  and Jinan Xu\textsuperscript{1}\thanks{ \ \ Jinan Xu is the corresponding author of the paper.} \\
  \textsuperscript{1}Beijing Jiaotong University, China \\
  \textsuperscript{2}Pattern Recognition Center, WeChat AI, Tencent Inc, China \\
  \texttt{adaxry@gmail.com} \\
  \texttt{\{fandongmeng, dayerzhang, withtomzhou\}@tencent.com} \\
  \texttt{\{chenyf, jaxu\}@bjtu.edu.cn} \\
}

\date{}

\begin{document}
\maketitle

\begin{abstract}
  Spoken Language Understanding (SLU) mainly involves two tasks, intent detection and slot filling, which are generally modeled jointly in existing works. 
  However, most existing models fail to fully utilize co-occurrence relations between slots and intents, which restricts their potential performance. To address this issue, in this paper we propose a novel $\mathbf{C}$ollaborative $\mathbf{M}$emory $\mathbf{N}$etwork (CM-Net) based on the well-designed block, named CM-block. The CM-block firstly captures slot-specific and intent-specific features from memories in a collaborative manner, and then uses these enriched features to enhance local context representations, based on which the sequential information flow leads to more specific (slot and intent) global utterance representations.
%   The CM-block firstly captures the slot-specific and intent-specific features from memories in a collaborative manner, and then uses these enriched features to enhance local context representations, finally the sequential information flow over local contexts leads to more specific (slot and intent) global representations in the utterance level.
  Through stacking multiple CM-blocks, our CM-Net is able to alternately perform information exchange among specific memories, local contexts and the global utterance, and thus incrementally enriches
  each other.
  We evaluate the CM-Net on two standard benchmarks (ATIS and SNIPS) and a self-collected corpus (CAIS).
  Experimental results show that the CM-Net achieves the state-of-the-art results on the ATIS and SNIPS in most of criteria, and significantly outperforms the baseline models on the CAIS.
  Additionally, we make the CAIS dataset publicly available for the research community \footnote{Code is available at:  https://github.com/Adaxry/CM-Net.}. 
\end{abstract}

%%%%%%%% Figure slot_intent_distribution %%%%%%%%
\begin{figure}[t!]
\begin{center}
     \scalebox{0.45}{
       \includegraphics[width=1\textwidth]{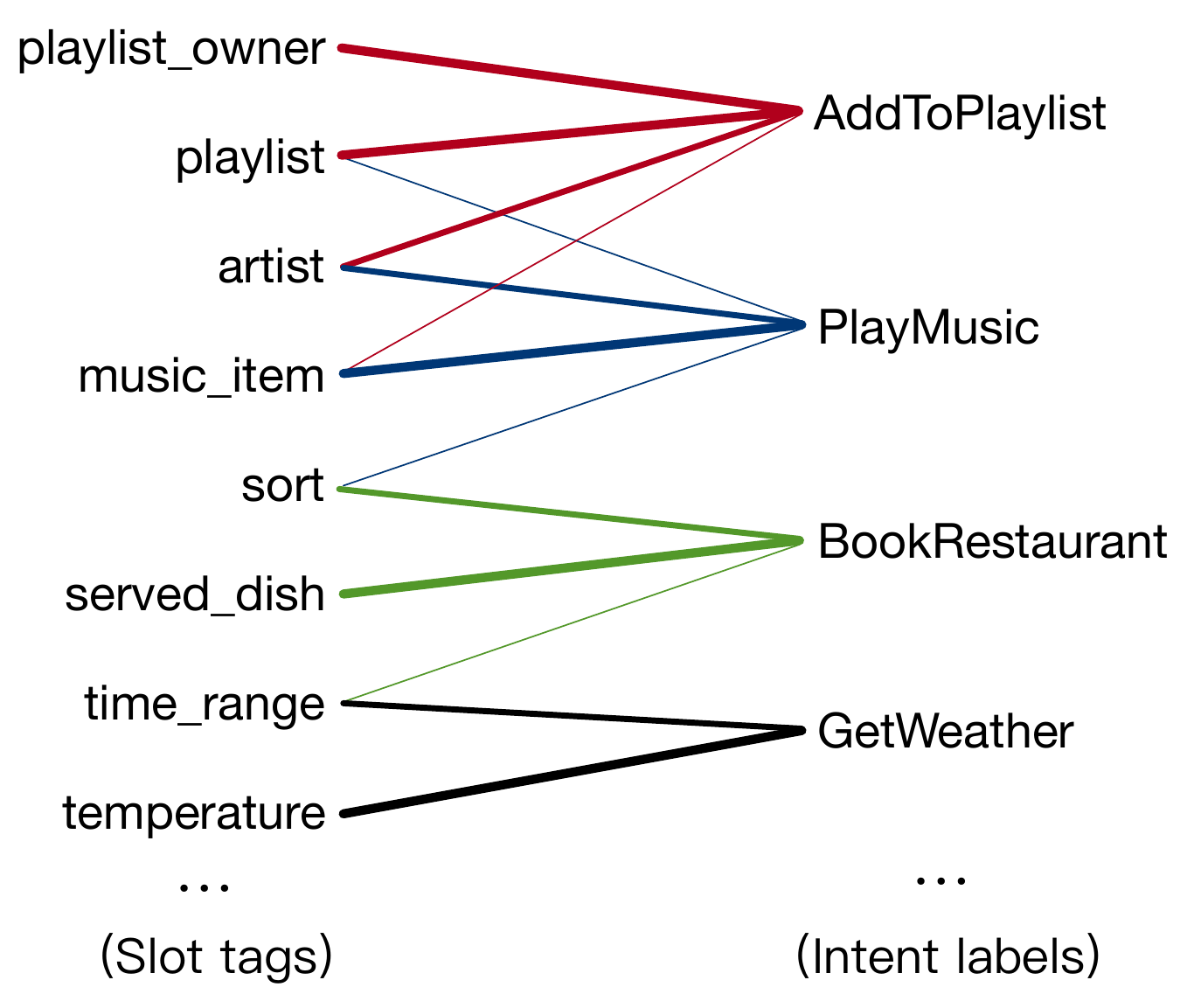}
      } \vspace{-6pt}
      \caption{
    %   Partial co-occurrence statistics
      Statistical association
      of slot tags (on the left) and intent labels (on the right) in the SNIPS, where colors indicate different intents and thicknesses of lines indicate proportions.
      } \vspace{-6pt}
      \label{slot_intent_distribution}  
 \end{center} 
\end{figure}

%%%%%%%  Table cases %%%%%%%%%
\begin{table*}[t!]
\begin{center}
\scalebox{1.0}{
\begin{tabular}{c l c c}
\hline \textbf{\#} & \textbf{Utterance} & \textbf{Slot tag} & \textbf{Intent}\\ \hline
1 & play \underline{Roy Orbison} tunes now & artist & PlayMusic \\
2 & add this \underline{Roy Orbison} song onto Women of Comedy & artist & AddToPlaylist \\
\hline
3 & book a spot for seven at a bar with chicken \underline{french} & served\_dish & BookRestaurant \\
%4 & book a table at a \underline{french} restaurant for fern and I & restaurant\_type & BookRestaurant \\
4 & book \underline{french} food for me and angeline at a restaurant & cuisine & BookRestaurant \\
\hline
\end{tabular}}
\end{center}
\caption{Examples in SNIPS with annotations of intent label for the utterance and slot tags for partial words.}
\label{cases}
\end{table*}

\section{Introduction}
Spoken Language Understanding (SLU) is a core component in dialogue systems. It typically aims to identify the intent and semantic constituents for a given utterance, which are referred as intent detection and slot filling, respectively. 
Past years have witnessed rapid developments in diverse deep learning models \cite{dl_2003,dl_2011} for SLU. To take full advantage of supervised signals of slots and intents, and share knowledge between them, most of existing works apply joint models that mainly based on CNNs \cite{cnn_joint_2013,simple_fast_2019}, RNNs \cite{recursive_joint_2014,attention_liu_2016}, and asynchronous bi-model \cite{bi_model_2018}.
Generally, these joint models encode words convolutionally or sequentially, and then aggregate hidden states into a utterance-level representation for the intent prediction, without interactions between representations of slots and intents.

Intuitively, slots and intents from similar fields tend to occur simultaneously, which can be observed from Figure \ref{slot_intent_distribution} and Table \ref{cases}. 
Therefore, it is beneficial to generate the representations of slots and intents with the guidance from each other.
Some works explore enhancing the slot filling task unidirectionally with the guidance from intent representations via gating mechanisms \cite{slot_gated_2018, self_attentive_2018}, while the predictions of intents lack the guidance from slots.
Moreover, the capsule network with dynamic routing algorithms \cite{capsule_2018} is proposed to perform interactions in both directions. However, there are still two limitations in this model. The one is that the information flows from words to slots, slots to intents and intents to words in a pipeline manner, which is to some extent limited in capturing complicated correlations among words, slots and intents. 
The other is that the local context information which has been shown highly useful for the slot filling \cite{local_window_2014}, is not explicitly modeled.
% Taking the utterances (row 1 and 2) in Table \ref{example} as examples, once the [words $|$ named entity] {\tt Roy Orbison} is recognized as {\tt artist}, the intent of the whole utterance is more likely to be {\tt PlayMusic} or {\tt AddToPlaylist}, instead of {\tt BookRestaurant} or {\tt GetWeather}. Similarly for the utterances (row 3, 4 and 5), the slot types of {\tt french} are more likely to be {\tt served\_dish}, {\tt restaurant\_type} or {\tt cuisine} instead of the [common $|$ general] {\tt country} when the intent is referred as {\tt BookRestaurant}.

In this paper, we try to address these issues, and thus propose a novel $\mathbf{C}$ollaborative $\mathbf{M}$emory $\mathbf{N}$etwork, named CM-Net. The main idea is to directly capture semantic relationships among words, slots and intents, which is conducted simultaneously at each word position in a collaborative manner. Specifically, we alternately perform information exchange among the task-specific features referred from memories, local context representations and global sequential information via the well-designed block, named CM-block, which consists of three computational components: 
\begin{itemize}
\item \textbf{Deliberate Attention}: 
Obtaining slot-specific and intent-specific representations from memories in a collaborative manner. 
% Conducting deliberate attentions over the slot and intent memories in a collaborative manner.  \vspace{-3pt}

% hidden states simultaneously
\item \textbf{Local Calculation}: 
Updating local context representations with the guidances of the referred slot and intent representations in the previous \emph{Deliberate Attention}. \vspace{-3pt}
% Performing information exchange among referred slot and intent representations, local ngrams to update word 
\item \textbf{Global Recurrence}: Generating specific (slot and intent) global sequential representations based on local context representations
from the previous \emph{Local Calculation}.
%\vspace{-3pt}
\end{itemize}
Above components in each CM-block are conducted consecutively, which are responsible for encoding information from different perspectives.
Finally, multiple CM-blocks are stacked together, and  construct our CM-Net. %, which enriches the information flows incrementally.

We firstly conduct experiments on two popular benchmarks, SNIPS \cite{snips_2018} and ATIS \cite{atis_1990,atis_2010}. Experimental results show that the CM-Net achieves the state-of-the-art results in 3 of 4 criteria ({\em e.g.,} intent detection accuracy on ATIS) on both benchmarks. Additionally, trials on our self-collected dataset, named CAIS, demonstrate the effectiveness and generalizability of the CM-Net. 

Our main contributions are as follows:
\begin{itemize}
\item We propose a novel CM-Net for SLU, which explicitly captures semantic correlations among words, slots and intents in a collaborative manner, and incrementally enriches the specific features, local context representations and global sequential representations through stacked CM-blocks.
\item Our CM-Net achieves the state-of-the-art results on two major SLU benchmarks (ATIS and SNIPS) in most of criteria.
\item We contribute a new corpus CAIS with manual annotations of slot tags and intent labels to the research community.
\end{itemize}

\section{Background}
\label{background}
In principle, the slot filling is treated as a sequence labeling task, and the intent detection is a classification problem. Formally, given an utterance $X = \{x_1, x_2, \cdots, x_N \}$ with $N$ words and its corresponding slot tags $Y^{slot} = \{y_1, y_2, \cdots, y_N \}$,
% with equal length,
the slot filling task aims to learn a parameterized mapping function $f_{\theta} : X \rightarrow Y $ from input words to slot tags. For the intent detection, it is designed to predict the intent label $\hat{y}^{int}$ for the entire utterance $X$ from the predefined label set $S^{int}$.

Typically, the input utterance is firstly encoded into a sequence of distributed representations $\mathbf{X} = \{\mathbf{x}_1, \mathbf{x}_2, \cdots, \mathbf{x}_N\}$ by character-aware and pre-trained word embeddings. Afterwards, the following bidirectional RNNs are applied to encode the embeddings $\mathbf{X}$ into context-sensitive representations $\mathbf{H} = \{\mathbf{h}_1, \mathbf{h}_2, \cdots, \mathbf{h}_N\}$. An external CRF \cite{CRF} layer is widely utilized to calculate conditional probabilities of slot tags:
\begin{equation}
        p (\mathbf{y}^{slot}|\mathbf{H}) = \frac{e^{F(\mathbf{H},\mathbf{y}^{slot})}}
        { \sum_{\widetilde{\mathbf{y}}^{slot} \in  \mathbf{Y}_x } e ^ { F (\mathbf{H}, \widetilde{\mathbf{y}}^{slot}) } } 
    \label{background_slot_crf}
\end{equation}
Here $\mathbf{Y}_x$ is the set of all possible sequences of tags, and $F(\cdot)$ is the score function calculated by:
\begin{equation}
        F(\mathbf{h},\mathbf{y}) = \sum_{i=1}^{N}{\mathbf{A}_{y_i, y_{i+1}}} + 
        \sum_{i=1}^{N}{\mathbf{P}_{i,y_i}}
        \label{background_crf_score}
\end{equation}
where $\mathbf{A}$ is the transition matrix that $\mathbf{A}_{i,j}$ indicates the score of a transition from $i$ to $j$, and $\mathbf{P}$ is the score matrix  output by RNNs. $P_{i,j}$ indicates the score of the $j^{th}$ tag of the $i^{th}$ word in a sentence \cite{lample_2016}.

When testing, the Viterbi algorithm \cite{viterbi} is used to search the sequence of slot tags with maximum score:
\begin{equation}
    \begin{split}
        & \hat{\mathbf{y}}^{slot} = \mathop{\arg\max} _ {\widetilde{\mathbf{y}}^{slot} \in \mathbf{Y}_x } F (\mathbf{H}, \widetilde{\mathbf{y}}^{slot} )
        \label{background_slot_pred}
    \end{split}
\end{equation}
As to the prediction of intent, the word-level hidden states $\mathbf{H}$ are firstly summarized into a utterance-level representation $\mathbf{v}^{int}$ via mean pooling (or max pooling or self-attention, {\em etc.}):
\begin{equation}
    \begin{split}
        & \mathbf{v}^{int} = \frac{1}{N} \sum_{i=1}^{N}{\mathbf{h}_t} 
    \end{split}
    \label{background_intent_rep}
\end{equation}
The most probable intent label $\hat{y}^{int}$ is predicted by softmax normalization over the intent label set:
\begin{equation}
    \begin{split}
        & \hat{y}^{int} = \mathop{\arg\max} _ {\widetilde{y} \in S^{int} } P (\widetilde{y} | \mathbf{v}^{int} ) \\
        & P(\widetilde{y} = j| \mathbf{v}^{int} ) = softmax(\mathbf{v}^{int})[j] 
    \end{split}
    \label{background_intent_pred}
\end{equation}

Generally, both tasks are trained jointly to minimize the sum of cross entropy from each individual task. Formally, the loss function of the join model is computed as follows:
\begin{equation}
    \begin{split}
    & L = (1 - \lambda) \cdot L^{slot} + \lambda \cdot L^{int} \\
    & L^{int} = - \sum\limits_{i=1}^{|S^{int}|}{\hat{y}^{int}_i} log(y^{int}_i) \\
    & L^{slot} = - \sum\limits_{j=1}^{N}\sum\limits_{i=1}^{|S^{slot}|}{\hat{y}^{slot}_{i,j}} log(y^{slot}_{i,j})
    \end{split}
\end{equation}
where $y^{int}_i$ and $y^{slot}_{i,j}$ are golden labels,
and $\lambda$ is hyperparameter,
% the weight to balance loss functions, 
% and $|S^{int}|$ and $|S^{int}|$  the size of intent label set, and
and $|S^{int}|$ is the size of intent label set,
and similarly for $|S^{slot}|$ .

%%%%%%%%%%%%%%  CM-Net Overview %%%%%%%%%%%%%%%%%%
\section{CM-Net}
\begin{figure}[t!]
\begin{center}
     \scalebox{0.48}{
       \includegraphics[width=1\textwidth]{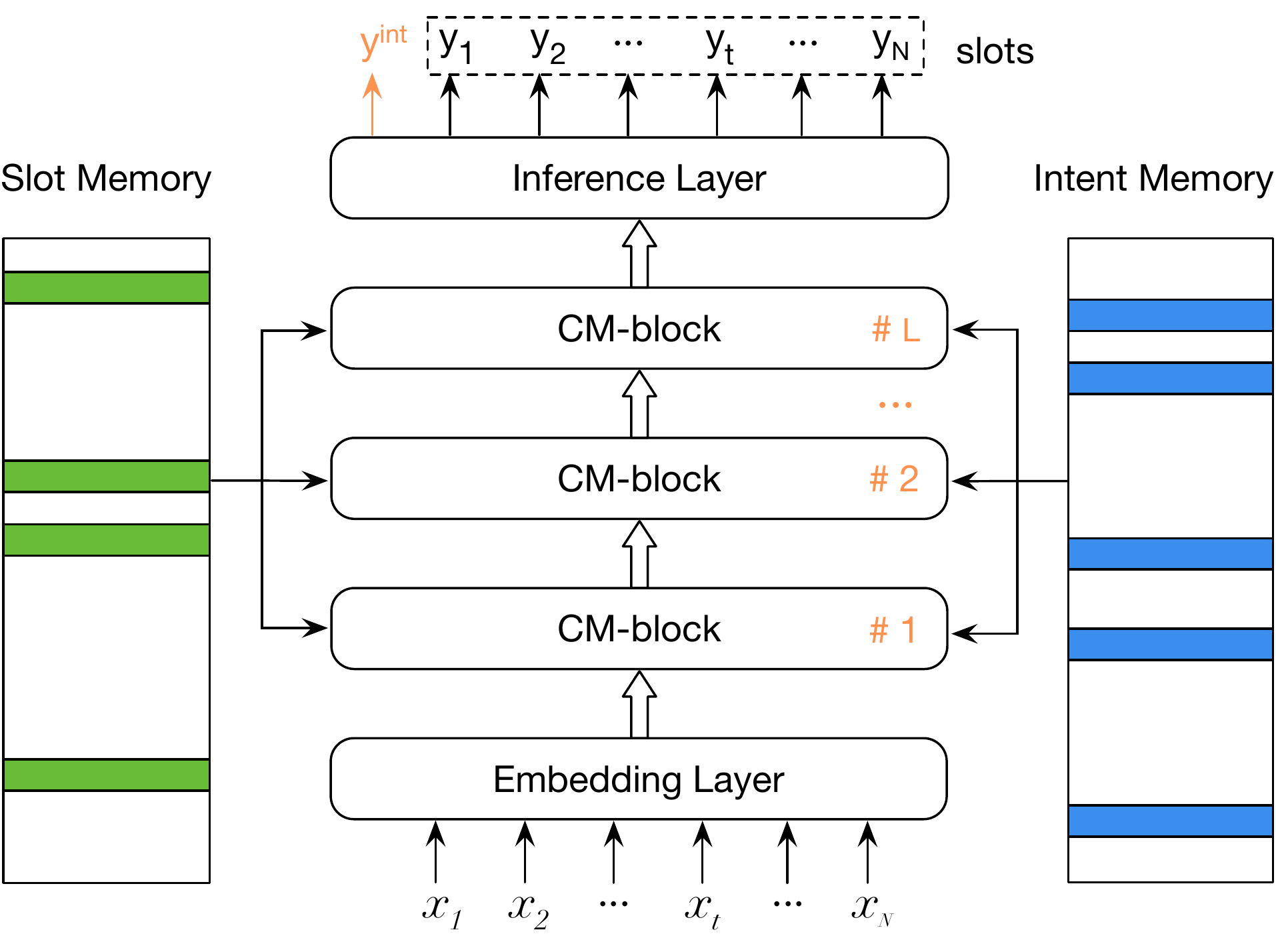}
      } \vspace{-19pt}
      \caption{Overview of our proposed CM-Net.
      The input utterance is firstly encoded with the Embedding Layer (bottom), and then is transformed by multiple CM-blocks with the assistance of both slot and intent memories (on both sides). Finally we make predictions of slots and the intent in the Inference Layer (top).
      }\vspace{-10pt}
      \label{overview}  
 \end{center}
\end{figure}

\subsection{Overview}
In this section, we start with a brief overview of our CM-Net and then proceed to introduce each module. As shown in Figure \ref{overview}, the input utterance is firstly encoded with the Embedding Layer, and then is transformed by multiple CM-blocks with the assistance of slot and intent memories, and finally make predictions in the Inference Layer.

\subsection{Embedding Layers}
\paragraph{Pre-trained Word Embedding}
The pre-trained word embeddings has been indicated as a de-facto standard of neural network architectures for various NLP tasks. 
We adapt the cased, 300d Glove\footnote{https://nlp.stanford.edu/projects/glove/} \cite{Glove} to initialize word embeddings, and keep them frozen.

\paragraph{Character-aware Word Embedding}
It has been demonstrated that character level information ({\em e.g.} capitalization and prefix) \cite{collobert_2011} is crucial for sequence labeling. We use one layer of CNN followed by max pooling to generate character-aware word embeddings.

\subsection{CM-block}
The CM-block is the core module of our CM-Net, which is designed with three computational components: \emph{Deliberate Attention}, \emph{Local Calculation} and \emph{Global Recurrence} respectively.

\subsubsection*{Deliberate Attention} To fully model semantic relations between slots and intents, 
we build the slot memory $\mathbf{M^{slot}} $ and  intent memory $\mathbf{M^{int}}$, and further devise a collaborative retrieval approach.
For the slot memory, it keeps $|S^{slot}|$ slot cells which are randomly initialized and updated as model parameters. Similarly for the intent memory.
At each word position, we take the hidden state $\mathbf{h}_t$ as query, and obtain slot feature $\mathbf{h}_t^{slot}$ and intent feature $\mathbf{h}_t^{int}$ from both memories by the deliberate attention mechanism, which will be illustrated in the following.

Specifically for the slot feature $\mathbf{h}_t^{slot}$, we firstly get a rough intent representation $\widetilde{\mathbf{h}}_t^{int}$ by the word-aware attention with hidden state $\mathbf{h}_t$ over the intent memory $\mathbf{M^{int}}$, and then obtain the final slot feature $\mathbf{h}_t^{slot}$ by the intent-aware attention over the slot memory $\mathbf{M^{slot}}$ with the intent-enhanced representation $[\mathbf{h}_t;\widetilde{\mathbf{h}}_t^{int}]$.
Formally, the above-mentioned procedures are computed as follows:
\begin{equation}
    \begin{split}
    \widetilde{\mathbf{h}}_t^{int} &= ATT(\mathbf{h_t}, \mathbf{M^{int}}) \\
    \mathbf{h}_t^{slot} &= ATT([\mathbf{h_t};\widetilde{\mathbf{h}}_t^{int}], \mathbf{M^{slot}}) 
    \end{split}
    \label{intent_aware_att}
\end{equation}
where $ATT(\cdot)$ is the query function calculated by the weighted sum of all cells $\mathbf{m}_i^{x}$ in memory $\mathbf{M}^{x}$ ($x \in \{slot, int\}$)  :
% by element-wise similarity:
\begin{equation}
    \begin{split}
    & ATT(\mathbf{h_t}, \mathbf{M}^{x}) = \sum\limits_{i}{\alpha_i \mathbf{m_i^{x}}} \\
    & \alpha_i = \frac{\exp ( \mathbf{u}^{\top}s_i )}
    {\sum_{j}{\exp ( \mathbf{u}^{\top}s_j) } } \\
    & s_i = \mathbf{h}_t^{\top} \mathbf{W} \mathbf{m_i^{x}}
    \end{split}
\end{equation}
Here $\mathbf{u}$ and $\mathbf{W}$ are model parameters. We name the above calculations of two-round attentions (Equation \ref{intent_aware_att}) as ``deliberate attention".

The intent representation $\mathbf{h}_t^{int}$ is computed by the deliberate attention as well:
% , as follows:
\begin{equation}
    \begin{split}
    \widetilde{\mathbf{h}}_t^{slot} & = ATT(\mathbf{h_t}, \mathbf{M^{slot}}) \\
    \mathbf{h}_t^{int} &=  ATT([\mathbf{h_t};\widetilde{\mathbf{h}}_t^{slot}], \mathbf{M^{int}})
    \end{split}
\end{equation}

These two deliberate attentions are conducted simultaneously at each word position in such collaborative manner, which guarantees adequate knowledge diffusions between slots and intents. The retrieved slot features $\mathbf{H}_t^{slot}$ and intent features $\mathbf{H}_t^{int}$ are utilized to provide guidances for the next local calculation layer. 

%%%%%%%     recurrent_details    %%%%%%%% 
\begin{figure}[t!]
\begin{center}
     \scalebox{0.45}{
       \includegraphics[width=1\textwidth]{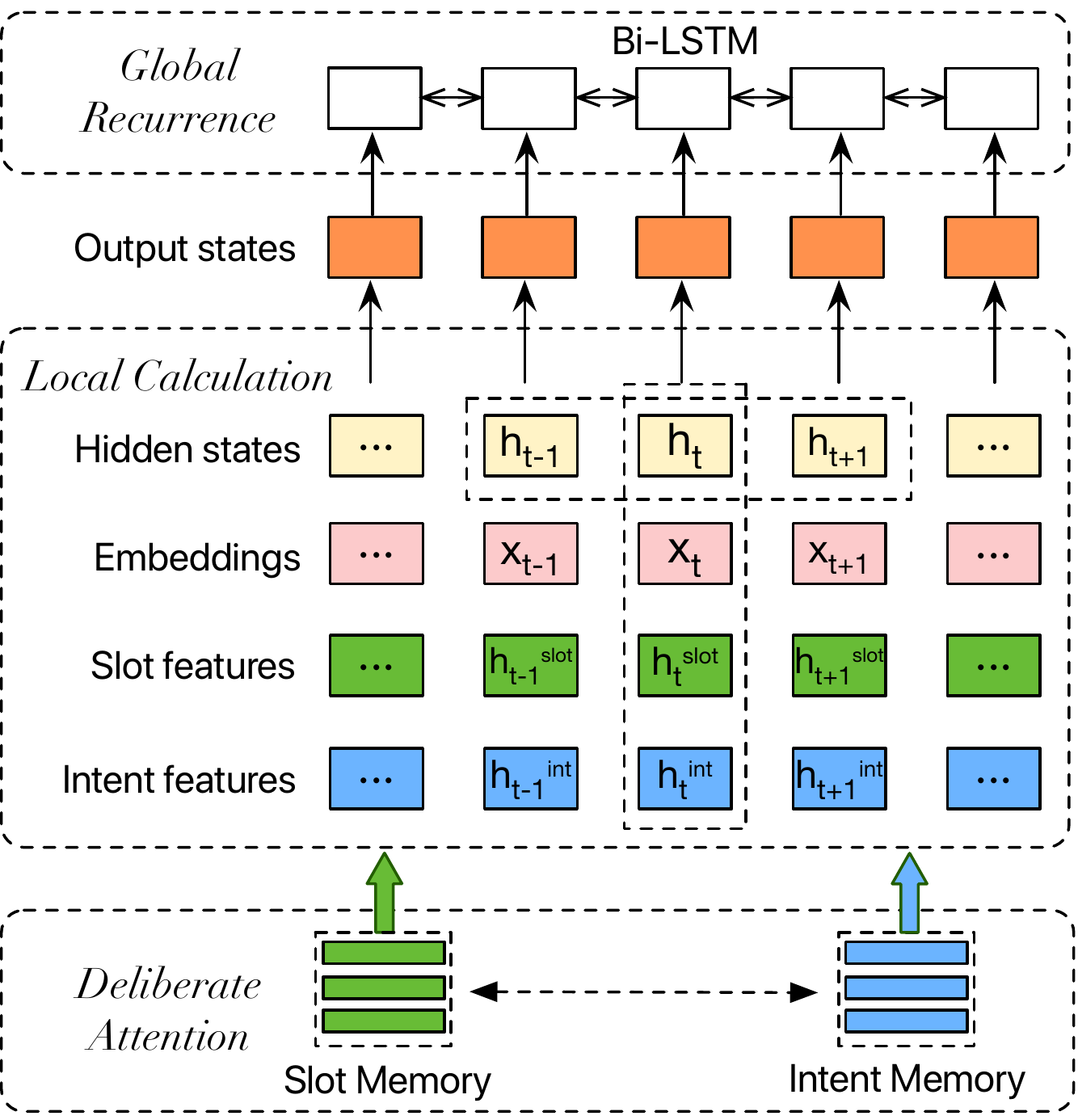}
      } \vspace{-5pt}
      \caption{
      The internal structure of our CM-Block,
      which is composed of \emph{deliberate attention}, \emph{local calculation} and  \emph{global recurrent} respectively.
      } 
      \label{recurrent_details}  
 \end{center}
\end{figure}

\subsubsection*{Local Calculation}
Local context information is highly useful for sequence modeling \cite{local_window_2016,local_window2_2016}. 
\citeauthor{SLSTM_2018} \shortcite{SLSTM_2018} propose the S-LSTM to encode both local and sentence-level information simultaneously, and it has been shown more powerful for text representation when compared with the conventional BiLSTMs.
% Inspired by the state transition procedure of the S-LSTM, 
We extend the S-LSTM with slot-specific features $\mathbf{H}_t^{slot}$ and intent-specific features $\mathbf{H}_t^{slot}$ retrieved from memories.
% and further supplement global sequential information in the next section, which is shown highly necessary and effective in our experiments.

Specifically, at each input position $t$, we take the local window context $\mathbf{\xi}_t$, word embedding $\mathbf{x}_t$, slot feature $\mathbf{h}_t^{slot}$  and intent feature $\mathbf{h}_t^{int}$ as inputs to conduct combinatorial calculation simultaneously.
Formally, in the $l^{th}$ layer, the hidden state $\mathbf{h_t}$ is updated as follows:
\begin{equation}
\begin{split}
  & \mathbf{\xi} _ { t } ^ { l-1 } = [ \mathbf{h} _ { t - 1 } ^ { l - 1 } , \mathbf{h} _ { t } ^ { l - 1 } , \mathbf{h} _ { t + 1 } ^ { l - 1 } ] 
 \\
 & \hat { \mathbf { i } } _ { t } ^ { l } =  \sigma ( \mathbf { W } _ { 1 }^{i} \mathbf { \xi } _ { t } ^ { l-1 } + \mathbf { W } _ { 2 }^{i} \mathbf { x } _ { t } + \mathbf { W } _ { 3 }^{i} \mathbf { h } ^ { slot } _ {t} +  \mathbf { W } _ { 4 }^{i} \mathbf { h } ^ { int } _ {t} ) 
 \\
 & \hat { \mathbf { o } } _ { t } ^ { l } =  \sigma ( \mathbf { W } _ { 1 }^{o} \mathbf { \xi } _ { t } ^ { l-1 } + \mathbf { W } _ { 2 }^{o} \mathbf { x } _ { t } + \mathbf { W } _ { 3 }^{o} \mathbf { h } ^ { slot } _ {t} +  \mathbf { W } _ { 4 }^{o} \mathbf { h } ^ { int } _ {t} ) 
 \\
 & \hat { \mathbf { f } } _ { t } ^ { l } =  \sigma ( \mathbf { W } _ { 1 }^{f} \mathbf { \xi } _ { t } ^ { l-1 } + \mathbf { W } _ { 2 }^{f} \mathbf { x } _ { t } + \mathbf { W } _ { 3 }^{f} \mathbf { h } ^ { slot } _ {t} +  \mathbf { W } _ { 4 }^{f} \mathbf { h } ^ { int } _ {t} ) 
  \\
  & \hat { \mathbf { l } } _ { t } ^ { l } =  \sigma ( \mathbf { W } _ { 1 }^{l} \mathbf { \xi } _ { t } ^ { l-1 } + \mathbf { W } _ { 2 }^{l} \mathbf { x } _ { t } + \mathbf { W } _ { 3 }^{l} \mathbf { h } ^ { slot } _ {t} +  \mathbf { W } _ { 4 }^{l} \mathbf { h } ^ { int } _ {t} ) 
  \\
  & \hat { \mathbf { r } } _ { t } ^ { l }  =  \sigma ( \mathbf { W } _ { 1 }^{r} \mathbf { \xi } _ { t } ^ { l-1 } + \mathbf { W } _ { 2 }^{r} \mathbf { x } _ { t } + \mathbf { W } _ { 3 }^{r} \mathbf { h } ^ { slot } _ {t} +  \mathbf { W } _ { 4 }^{r} \mathbf { h } ^ { int } _ {t} ) 
  \\
  & \mathbf { u } _ { t } ^ { l }  =  \tanh ( \mathbf { W } _ { 1 }^{u} \mathbf { \xi } _ { t } ^ { l-1 } + \mathbf { W } _ { 2 }^{u} \mathbf { x } _ { t } + \mathbf { W } _ { 3 }^{u} \mathbf { h } ^ { slot } _ {t} +  \mathbf { W } _ { 4 }^{u} \mathbf { h } ^ { int } _ {t} ) \\
  & \mathbf{i}_t^l, \mathbf{f}_t^l, \mathbf{l}_t^l, \mathbf{r}_t^l = softmax(\hat{\mathbf{i}}_t^l, \hat{\mathbf{f}}_t^l, \hat{\mathbf{l}}_t^l, \hat{\mathbf{r}}_t^l) \\
  & \mathbf{c}_t^l = \mathbf{f}_t^l \odot \mathbf{c}_{t}^{l-1} + \mathbf{l}_t^l \odot \mathbf{c}_{t-1}^{l-1} + \mathbf{r}_t^l \odot \mathbf{c}_{t+1}^{l-1} + \mathbf{i}_t^l \odot \mathbf{u}_{t}^{l-1} \\
  & \mathbf{h}_t^l = \mathbf{o}_t^l \odot \tanh{\mathbf{c}_t^l}
  \end{split}
\end{equation}
where $\mathbf { \xi } _ { t } ^ { l }$ is the concatenation of hidden states in a local window, and $\mathbf{i}_t^l$, $\mathbf{f}_t^l$, $\mathbf{o}_t^l$, $\mathbf{l}_t^l$ and $\mathbf{r}_t^l$ are gates to control information flows, and $\mathbf{W}_n^x$ $(x \in \{ i, o, f, l, r, u\}, n \in \{1, 2, 3, 4\})$ are model parameters. More details about the state transition can be referred in \cite{SLSTM_2018}. In the first CM-block, the hidden state $\mathbf{h}_t$ is initialized with the corresponding word embedding. In other CM-blocks, the $\mathbf{h}_t$ is inherited from the output of the adjacent lower CM-block.

At each word position of above procedures, the hidden state is updated with abundant information from different perspectives, namely word embeddings, local contexts, slots and intents representations. The local calculation layer in each CM-block has been shown highly useful for both tasks, and especially for the slot filling task, which will be validated in our experiments in Section \ref{ablation_experiment}.

\subsubsection*{Global Recurrence}
Bi-directional RNNs, especially the BiLSTMs \cite{LSTM_1997} are regarded to encode both past and future information of a sentence, which have become a dominant method in various sequence modeling tasks \cite{first_LSTM_2003,lstm_lm_2012}. 
The inherent nature of BiLSTMs is able to supplement  global sequential information, which is insufficiently modeled in the previous local calculation layer. 
Thus we apply an additional BiLSTMs layer upon the local calculation layer in each CM-block. By taking the slot- and intent-specific local context representations as inputs, we can obtain more specific global sequential representations. Formally, it takes the hidden state $\mathbf{h}_t^{l-1}$ inherited from the local calculation layer as input, and conduct recurrent steps as follows:
\begin{equation}
    \begin{split}
    & \mathbf{h}_t^{l} = [\overrightarrow{\mathbf{h}_t^{l}}; \overleftarrow{\mathbf{h}_t}^{l}] \\
    & \overrightarrow{\mathbf{h}_t^{l}} = \overrightarrow{\mathbf{LSTM}}(\mathbf{h}_t^{l-1}, \overrightarrow{\mathbf{h}}_{t-1}^{l}; \mathbf{\overrightarrow{\theta}}) \\
    & \overleftarrow{\mathbf{h}_t^{l}} = \overleftarrow{\mathbf{LSTM}}(\mathbf{h}_t^{l-1}, \overleftarrow{\mathbf{h}}_{t+1}^{l}; \mathbf{\overleftarrow{\theta}})
    \end{split}
\end{equation}
The output ``states" of the BiLSTMs are taken as ``states" input of the local calculation in next CM-block. The global sequential information encoded by the BiLSTMs is shown necessary and effective for both tasks in our experiments in Section  \ref{ablation_experiment}.

\subsection{Inference Layer}
After multiple rounds of interactions among local context representations, global sequential information, slot and intent features, we conduct predictions upon the final CM-block.
For the predictions of slots, we take the hidden states $\mathbf{H}$ along with the retrieved slot $\mathbf{H}^{slot}$ representations (both are from the final CM-block) as input features, and then conduct predictions of slots similarly with the Equation (\ref{background_slot_pred}) in Section \ref{background}:
\begin{equation}
    \begin{split}
        & \hat{\mathbf{y}}^{slot} = \mathop{\arg\max} _ {\widetilde{\mathbf{y}}^{slot} \in \mathbf{Y}_x } F ([\mathbf{H}; \mathbf{H}^{slot}], \widetilde{\mathbf{y}}^{slot} )
    \end{split}
    \label{slot_pred}
\end{equation}
For the prediction of intent label, we firstly aggregate the hidden state $\mathbf{h}_t$ and the retrieved intent representation $\mathbf{h}_t^{int}$ at each word position (from the final CM-block as well) via mean pooling:
\begin{equation}
    \begin{split}
        \mathbf{v}^{int} = \frac{1}{N} \sum\limits_{t}^{N}{[ \mathbf{h_t}};\mathbf{h}_t^{int}]
    \end{split}
    \label{intent_concat}
\end{equation}
and then take the summarized vector $\mathbf{v}^{int}$ as input feature to conduct prediction of intent consistently with the Equation (\ref{background_intent_pred}) in Section \ref{background}.

%%%%%%%%%%%%% Table dataset %%%%%%%%%%%
\begin{table}[t!]
\begin{center}
\scalebox{0.9}{
\begin{tabular}{l|c|c|c}
\hline
\textbf{Dataset} & \textbf{SNIPS} & \textbf{ATIS} & \textbf{CAIS} \\
\hline
Vocab Size  & 11241  & 722 & 2146 \\
Average Length  & 9.15  & 11.28  & 8.65 \\
\# Intents  & 7 & 18  & 11 \\
\# Slots  & 72 & 128  & 75 \\
\# Train Set & 13084  & 4478  & 7995 \\
\# Validation Set  & 700  & 500  & 994 \\
\# Test Set & 700  & 893  & 1012 \\
\hline
\end{tabular}}
\end{center} \vspace{-5pt}
\caption{Dataset statistics.} \vspace{-5pt}
\label{data_statistics}
\end{table}

%%%%%%%%%%%%% Table result %%%%%%%%%%
\begin{table*}[t!]
\begin{center}
\scalebox{0.9}{
\begin{tabular}{l|c|c|c|c}
\hline \multirow{2}*{Models} & \multicolumn{2}{c}{\textbf{SNIPS}} & \multicolumn{2}{|c}{\textbf{ATIS}} \\
\cline{2-5}
~ & Slot ($F_1$) & Intent ($Acc$) & Slot ($F_1$) & Intent ($Acc$) \\
\hline
Joint GRU \cite{a_joint_2016} & -- & -- & 95.49 & 98.10 \\
Self-Attention, Intent Gate\cite{self_attentive_2018} & -- & -- & 96.52 & 98.77  \\
Bi-model \cite{bi_model_2018} & -- & -- & \textbf{96.89} & 98.99  \\
Attention Bi-RNN \cite{attention_liu_2016} * & 87.80 & 96.70 & 95.98 & 98.21 \\
Joint Seq2Seq \cite{seq2seq_2016} * & 87.30 & 96.90 & 94.20 & 92.60  \\
Slot-Gated (Intent Atten.) \cite{slot_gated_2018} & 88.30 & 96.80 & 95.20 & 94.10 \\
Slot-Gated (Full Atten.) \cite{slot_gated_2018} & 88.80 & 97.00 & 94.80 & 93.60 \\
CAPSULE-NLU\cite{capsule_2018}& 91.80 & 97.70 & 95.20 & 95.00 \\
Dilated CNN, Label-Recurrent \cite{simple_fast_2019} & 93.11 & 98.29 & 95.54 & 98.10 \\
Sentence-State LSTM \cite{SLSTM_2018} $\dagger$ & 95.80 &  98.30 & 95.65 & 98.21 \\
BiLSTMs + EMLoL \cite{EMLo_bilstm_crf_2018} &  93.29 & 98.83 & 95.62 & 97.42 \\
BiLSTMs + EMLo \cite{EMLo_bilstm_crf_2018} &  93.90 & \textbf{99.29}  & 95.42 & 97.30 \\
Joint BERT \cite{bert_2019} & 97.00 & 98.60 & 96.10 & 97.50 \\
\hline
CM-Net (Ours) & \textbf{97.15} & \textbf{99.29} & 96.20 & \textbf{99.10}  \\
\hline
\end{tabular}}
\end{center}\vspace{-3pt}
\caption{Results on test sets of the SNIPS and ATIS, where our CM-Net achieves state-of-the-art performances in most cases. ``*" indicates that results are retrieved from Slot-Gated \cite{slot_gated_2018}, and  ``$\dagger$" indicates our implementation.} \vspace{-3pt}
\label{main_result} 
\end{table*}

\section{Experiments}
\label{experiments}

\subsection{Datasets and Metrics}
\label{datasets}
We evaluate our proposed CM-Net on three real-word datasets, and statistics are listed in Table \ref{data_statistics}.

\paragraph{ATIS} The Airline Travel Information Systems (ATIS) corpus \cite{atis_1990} is the most widely used benchmark for the SLU research. 
Please note that, there are extra named entity features in the ATIS, which almost determine slot tags. These hand-crafted features are not generally available in open domains \cite{a_joint_2016,a_joint_2014}, therefore we train our model purely on the training set without additional hand-crafted features.

\paragraph{SNIPS} SNIPS Natural Language Understanding benchmark \footnote{https://github.com/snipsco/nlu-benchmark/tree/master/2017-06-custom-intent-engines} \cite{snips_2018} is collected in a crowsourced fashion by Snips. The intents of this dataset are more balanced when compared with the ATIS. We split another 700 utterances for validation set following previous works \cite{slot_gated_2018, capsule_2018}.

\paragraph{CAIS} 
We collect utterances from the $\mathbf{C}$hinese $\mathbf{A}$rtificial $\mathbf{I}$ntelligence $\mathbf{S}$peakers (CAIS), and annotate them with slot tags and intent labels.
The training, validation and test sets are split by the distribution of intents, where detailed statistics are provided in the supplementary material. Since the utterances are collected from speaker systems in the real world, intent labels are partial to the \emph{PlayMusic} option.
We adopt the BIOES tagging scheme for slots instead of the BIO2 used in the ATIS, since previous studies have highlighted meaningful improvements with this scheme \cite{BIES} in the sequence labeling field. 

\paragraph{Metrics} 
Slot filling is typically treated as a sequence labeling problem, and thus we take the conlleval \footnote{https://www.clips.uantwerpen.be/conll2000/chunking/\\conlleval.txt} as the token-level $F_1$  metric. The intent detection is evaluated with the classification accuracy. Specially, several utterances in the ATIS are tagged with more than one labels. Following previous works \cite{atis_2010,a_joint_2016}, we count an utterrance as a correct classification if any ground truth label is predicted.

\subsection{Implementation Details}

All trainable parameters in our model are initialized by the method described in \citeauthor{Xavier} \shortcite{Xavier}. We apply dropout \cite{dropout} to the embedding layer and hidden states with a rate of 0.5. All models are optimized by the Adam optimizer \cite{Adam} with gradient clipping of 3 \cite{gradient_clip}. 
The initial learning rate $\alpha$ is set to 0.001, and decrease with the growth of training steps. We monitor the training process on the validation set and report the final result on the test set. 
One layer CNN with a filter of size 3 and max pooling are utilized to generate 100d word embeddings. The cased 300d Glove is adapted to initialize word embeddings, and kept fixed when training. In auxiliary experiments, the output hidden states of BERT are taken as additional word embeddings and kept fixed as well. We share parameters of both memories with the parameter matrices in the corresponding softmax layers, which can be taken as introducing supervised signals into the memories to some extent. We conduct hyper-parameters tuning for layer size (finally set to 3) and loss weight $\lambda$ (finally set to 0.5), and empirically set other parameters to the values listed in the supplementary material.

\subsection{Main Results}
Main results of our CM-Net on the SNIPS and ATIS are shown in Table \ref{main_result}. Our CM-Net achieves the state-of-the-art results on both datasets in terms of slot filling $F_1$ score and intent detection accuracy, except for the $F_1$ score on the ATIS. We conjecture that the named entity feature in the ATIS has a great impact on the slot filling result as illustrated in Section \ref{datasets}. Since the SNIPS is collected from multiple domains with more balanced labels when compared with the ATIS, the slot filling $F_1$ score on the SNIPS is able to demonstrate the superiority of our CM-Net.

It is noteworthy that the CM-Net achieves comparable results when compared with models that exploit additional language models \cite{EMLo_bilstm_crf_2018,bert_2019}. We conduct auxiliary experiments by leveraging the well-known BERT \cite{BERT} as an external resource for a relatively fair comparison with those models, and report details in Section \ref{effects_of_lm}.

\begin{figure}[t!]
\begin{center}
     \scalebox{0.48}{
       \includegraphics[width=1\textwidth]{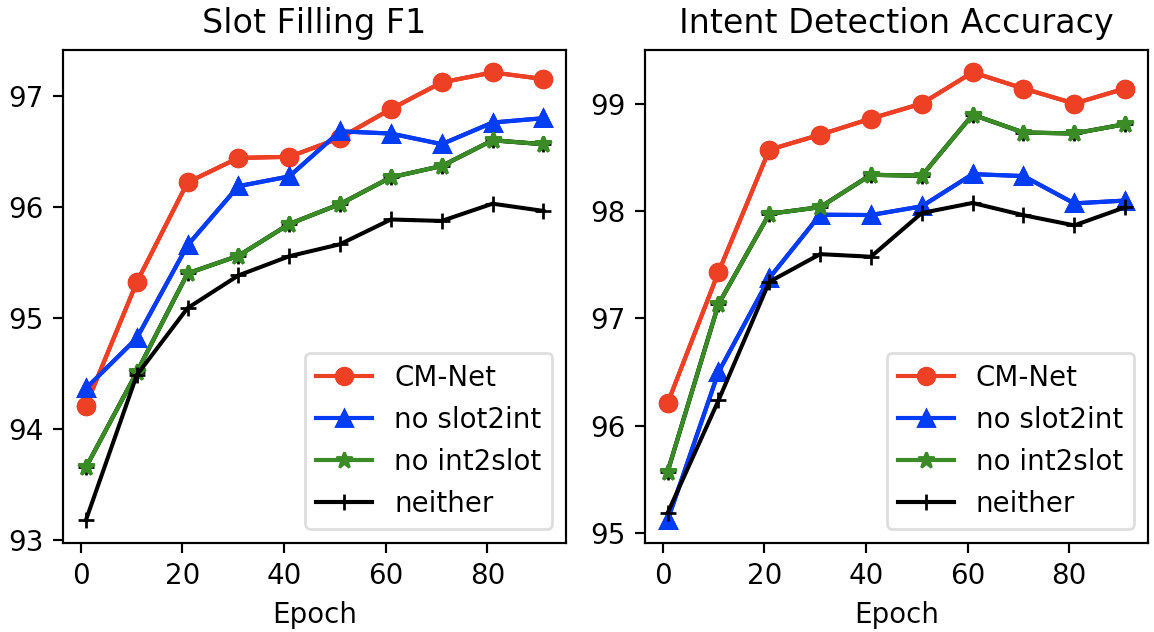}
      } 
      \caption{Investigations of the collaborative retrieval approach on slot filling (on the left) and intent detection (on the right), where ``no slot2int" indicates removing slow-aware attention for the intent representation, and similarly for ``no int2slot" and ``neither".
    %   means a plain attention mechanism with two completely independent memories.
      } 
      \label{analysis_mem}  
 \end{center}
\end{figure}

%%%%%%%%%%%%%%%% Table ablation %%%%%%%%%
\begin{table}[t!]
\begin{center}
\scalebox{0.9}{
\begin{tabular}{c|l|c|c}
\hline \multirow{2}*{\textbf{\#}} & \multirow{2}*{Models} & \multicolumn{2}{c}{\textbf{SNIPS}}  \\
\cline{3-4}
~ & ~ & Slot ($F_1$) & Intent ($Acc$)\\
\hline
0 & CM-Net &  \textbf{97.15} & \textbf{99.29} \\
1 & -- slot memory & 96.64  & 99.14 \\
2 & -- intent memory & 96.95 & 98.84 \\
3 & -- local calculation & 96.73 & 99.00 \\
4 & -- global recurrence & 96.80 & 98.57 \\
\hline
\end{tabular}}
\end{center}
\caption{Ablation experiments on the SNIPS to investigate the impacts of various components, where ``- slot memory" indicates removing the slot memory and its interactions with other components correspondingly. Similarly for the other options.}
\label{ablation_studies}
\end{table}

\section{Analysis}
Since the SNIPS corpus is collected from multiple domains and its label distributions are more balanced when compared with the ATIS, we choose the SNIPS to elucidate properties of our CM-Net and conduct several additional experiments.

\subsection{Whether Memories Promote Each Other?}
\label{analysis_promote_each_other}
In the CM-Net, the deliberate attention mechanism is proposed in a collaborative manner to perform information exchange between slots and intents. We conduct experiments to verify whether such kind of knowledge diffusion in both memories can promote each other. 
More specifically, we remove one unidirectional diffusion ({\em e.g.} from slot to intent) or both in each experimental setup. The results are illustrated in Figure \ref{analysis_mem}.

We can observe obvious drops on both tasks when both directional knowledge diffusions are removed (CM-Net {\em vs.} neither). For the slot filling task (left part in Figure \ref{analysis_mem}), the $F_1$ scores decrease slightly when the knowledge from slot to intent is blocked (CM-Net {\em vs.} ``no slot2int"), and a more evident drop occurs when the knowledge from intent to slot is blocked (CM-Net {\em vs.} ``no int2slot"). Similar observations can be found for the intent detection task (right part in Figure \ref{analysis_mem}).

In conclusion, the bidirectional knowledge diffusion between slots and intents are necessary and effective to promote each other.

\subsection{Ablation Experiments}
\label{ablation_experiment}
We conduct ablation experiments to investigate the impacts of various components in our CM-Net. In particular, we remove one component among slot memory, intent memory, local calculation and global recurrence. Results of different combinations are presented in Table \ref{ablation_studies}. 

Once the slot memory and its corresponding interactions with other components are removed, scores on both tasks decrease to some extent, and a more obvious decline occurs for the slot filling (row 1 {\em vs.} row 0), which is consistent with the conclusion of Section \ref{analysis_promote_each_other}. Similar observations can be found for the intent memory (row 2). The local calculation layer is designed to capture better local context representations, which has an evident impact on the slot filling and slighter effect on the intent detection (row 3 {\em vs.} row 0). Opposite observations occur in term of global recurrence, which is supposed to model global sequential information and thus has larger effect on the intent detection (row 4 {\em vs.} row 0).

%%%%%%%%%%%%%%% Table LM %%%%%%%%%%
\begin{table}[t!]
\begin{center}
\scalebox{0.9}{
\begin{tabular}{l|c|c}
\hline \multirow{2}*{Models} & \multicolumn{2}{c}{\textbf{SNIPS}}  \\
\cline{2-3}
~ & Slot ($F_1$) & Intent ($Acc$)\\
\hline
BiLSTMs + EMLoL  &  93.29 & 98.83 \\
BiLSTMs + EMLo &  93.90 & 99.29 \\
Joint BERT & 97.00 & 98.60 \\
\hline
CM-Net + BERT & \textbf{97.31} & \textbf{99.32}  \\
\hline
\end{tabular}}
\end{center}
\caption{Results on the SNIPS benchmark with the assistance of pre-trained language model, where we establish new state-of-the-art results on the SNIPS.}
\label{lm_result}
\end{table}

\subsection{Effects of Pre-trained Language Models}
\label{effects_of_lm}
Recently, there has been a growing body of works exploring neural language models that trained on massive corpora to learn contextual representations ({\em e.g.} BERT \shortcite{BERT} and EMLo \shortcite{EMLo}).
% in both character level (Flair \citeyear{Flair_18}, EMLo \citeyear{EMLo}) and token level (BERT \citeyear{BERT}). 
Inspired by the effectiveness of language model embeddings, we conduct experiments by leveraging the BERT as an additional feature. The results emerged in Table \ref{lm_result} show that we establish new state-of-the-art results on both tasks of the SNIPS. 

\subsection{Evaluation on the CAIS}
We conduct experiments on our self-collected CAIS to evaluate the generalizability in different language.  We apply two baseline models for comparison, one is the popular \emph{BiLSTMs + CRF} architecture \cite{BLSTM_CRF_2015} for sequence labeling task, and the other one is the more powerful sententce-state LSTM \cite{SLSTM_2018}. The results listed in Table \ref{self_result} demonstrate the generalizability and effectiveness of our CM-Net when handling various domains and different languages.

%%%%%%%%%%%%% Table self %%%%%%%%%%%%
\begin{table}[t!]
\begin{center}
\scalebox{0.9}{
\begin{tabular}{l|c|c}
\hline \multirow{2}*{ Models } & \multicolumn{2}{c}{ \textbf{CAIS} }  \\
\cline{2-3}
~ & Slot ($F_1$) & Intent ($Acc$)\\
\hline
BiLSTMs + CRF &  85.32 & 93.25 \\
S-LSTM + CRF $\dagger$ &  85.74 & 94.36 \\
CM-Net & \textbf{86.16} & \textbf{94.56}  \\
\hline
\end{tabular}}
\end{center}
\caption{Results on our CAIS dataset, where ``$\dagger$" indicates our implementation of the S-LSTM.}
\label{self_result}
\end{table}

\section{Related Work}
\paragraph{Memory Network} 
Memory network is a general machine learning framework introduced by \citeauthor{memory_2014} \shortcite{memory_2014}, which have been shown effective in question answering \cite{memory_2014,memory_2015}, machine translation \cite{memory_mt_2016,memory_mt_2017}, aspect level sentiment classification \cite{memory_absa_2016},  {\em etc.} For spoken language understanding, \citeauthor{memory_slu_2016} \shortcite{memory_slu_2016} introduce memory mechanisms to encode historical utterances. In this paper, we propose two memories to explicitly capture the semantic correlations between slots and the intent in a given utterance, and devise a novel collaborative retrieval approach.

\paragraph{Interactions between slots and intents}
Considering the semantic proximity between slots and intents, some works propose to enhance the slot filling task unidirectionally with the guidance of intent representations via gating mechanisms \cite{slot_gated_2018, self_attentive_2018}. Intuitively, the slot representations are also instructive to the intent detection task and thus bidirectional interactions between slots and intents are benefical for each other.
\citeauthor{capsule_2018} \shortcite{capsule_2018} propose a hierarchical capsule network to perform interactions from words to slots, slots to intents and intents to words in a pipeline manner, which is relatively limited in capturing the complicated correlations among them. In our CM-Net, information exchanges are performed simultaneously with knowledge diffusions in both directions. 
% Furthermore, the local calculation component in our CM-Net can capturing better local context representations.
The experiments demonstrate the superiority of our CM-Net in capturing the semantic correlations between slots and intents.

\paragraph{Sentence-State LSTM}
\citeauthor{SLSTM_2018} \citeyear{SLSTM_2018} propose a novel graph RNN named S-LSTM, which models sentence between words simultaneously. Inspired by the new perspective of state transition in the S-LSTM, we further extend it with task-specific ({\em i.e., } slots and intents) representations via our collaborative memories. In addition, the global information in S-LSTM is modeled by aggregating the local features with gating mechanisms, which may lose sight of sequential information of the whole sentence. Therefore, We apply external BiLSTMs to supply global sequential features, which is shown highly necessary for both tasks in our experiments.

\section{Conclusion}
We propose a novel $\mathbf{C}$ollaborative $\mathbf{M}$emory $\mathbf{N}$etwork (CM-Net) for jointly modeling slot filling and intent detection. The CM-Net is able to explicitly capture the semantic correlations among words, slots and intents in a collaborative manner, and incrementally enrich the information flows with local context and global sequential information.
% We propose a novel $\mathbf{C}$ollaborative $\mathbf{M}$emory $\mathbf{N}$etwork (CM-Net) for jointly modeling slot filling and intent detection. The CM-Net is composed of multiple well-design CM-blocks, which firstly model the semantic correlations among words,  slots and intents in a collaborative manner, and then capture local context representations simultaneously, and finally encode global sequential information.  
Experiments on two standard benchmarks and our CAIS corpus demonstrate the effectiveness and generalizability of our proposed CM-Net. In addition, we contribute the new corpus (CAIS) to the research community.

% In the future, we will investigate how to import unsupervised knowledge into the CM-Net, and explore its effectiveness on more languages.

\section*{Acknowledgments}
Liu, Chen and Xu are supported by the National Natural Science Foundation of China (Contract 61370130, 61976015, 61473294 and 61876198), and the Beijing Municipal Natural Science Foundation (Contract 4172047), and the International Science and Technology Cooperation Program of the Ministry of Science and Technology (K11F100010). We sincerely thank the anonymous reviewers for their thorough reviewing and valuable suggestions.

\bibliography{emnlp-ijcnlp-2019}
\bibliographystyle{acl_natbib}

\appendix

\end{document}